\documentclass[sigconf]{acmart}
\AtBeginDocument{%
  }

\copyrightyear{2026}
\acmYear{2026}
\setcopyright{cc}
\setcctype{by}
\acmConference[MM '26]{Proceedings of the 34th ACM International Conference on Multimedia}{November 10--14, 2026}{Rio de Janeiro, Brazil}
\acmBooktitle{Proceedings of the 34th ACM International Conference on Multimedia (MM '26), November 10--14, 2026, Rio de Janeiro, Brazil}
\acmDOI{10.1145/3767308.3835827}
\acmISBN{979-8-4007-2213-4/2026/11}
\usepackage{xcolor}
\usepackage{fontawesome}
\usepackage{colortbl}
\usepackage{booktabs}
\usepackage{graphicx}
\usepackage[many]{tcolorbox}
\usepackage{caption}
\usepackage{stackengine}
\usepackage{multirow}
\usepackage{afterpage}
\usepackage{subcaption}
\usepackage[ruled,linesnumbered]{algorithm2e}
\definecolor{userblue}{RGB}{235, 243, 255}
\definecolor{userborder}{RGB}{100, 150, 250}
\definecolor{agentgreen}{RGB}{240, 250, 240}
\definecolor{agentborder}{RGB}{120, 180, 120}
\definecolor{toolorange}{RGB}{255, 245, 235}
\definecolor{toolborder}{RGB}{250, 180, 100}

\tcbset{
    chatbox/.style={
        width=\linewidth,
        boxrule=0.5mm,
        arc=2mm,
        fonttitle=\bfseries,
        left=2mm, right=2mm, top=2mm, bottom=2mm
    }
}
\newcommand{\methodwithconf}[2]{%
    #1\rlap{\kern0.2em\raisebox{-0.2ex}{\scriptsize ~#2}}%
}

\newcommand{\methodwithconfdag}[2]{%
    #1$^{\dagger}$\rlap{\kern-1.2em\raisebox{-0.8ex}{\scriptsize ~#2}}%
}

\begin{document}

\title{PSGAgent: Enhancing MLLMs' Dynamic Visual Perception for Panoptic Scene Graph Generation via Agentic Reasoning}

\author{Qingyun Liu}
\email{liuqingyun25@stu.pku.edu.cn}
\orcid{0009-0004-0861-2762}
\affiliation{%
  \institution{Peking University}
  \city{Beijing}
  \country{China}
}

\author{Yuankun Liu}
\email{liuyk@stu.pku.edu.cn}
\orcid{0009-0009-3600-1586}
\affiliation{%
  \institution{Peking University}
  \city{Beijing}
  \country{China}
}

\author{Jiaye Feng}
\email{jyfeng25@stu.pku.edu.cn}
\orcid{0009-0009-7163-6827}
\affiliation{%
  \institution{Peking University}
  \city{Beijing}
  \country{China}
}

\author{Jie Huang}
\email{huangjie25@stu.pku.edu.cn}
\orcid{0009-0000-9015-2582}
\affiliation{%
  \institution{Peking University}
  \city{Beijing}
  \country{China}
}

\author{Zonghong Dai}
\email{zhdai23@m.fudan.edu.cn}
\orcid{0009-0006-7723-4130}
\affiliation{%
  \institution{AlignBase}
  \city{Beijing}
  \country{China}
}

\author{Haochen Li}
\email{lihaochen@jidianqiyuan.com}
\orcid{0000-0002-3197-4073}
\affiliation{%
  \institution{AlignBase}
  \city{Beijing}
  \country{China}
}

\author{Jingkun Wang}
\email{wjk@pku.edu.cn}
\orcid{0000-0002-8658-9690}
\affiliation{%
  \institution{AlignBase}
  \city{Beijing}
  \country{China}
}

\author{Weiping Li}
\email{wpli@ss.pku.edu.cn}
\orcid{0000-0003-2958-3097}
\affiliation{%
  \institution{Peking University}
  \city{Beijing}
  \country{China}
}

\author{Tong Mo}
\correspondingauthor
\email{motong@ss.pku.edu.cn}
\orcid{0000-0002-3564-4610}
\affiliation{%
  \institution{Peking University}
  \city{Beijing}
  \country{China}
}

\renewcommand{\shortauthors}{Qingyun Liu et al.}
\begin{abstract}
Panoptic Scene Graph generation (PSG) requires holistic yet fine-grained reasoning over complex relations among objects.
However, current Multimodal Large Language Models (MLLMs) applied to PSG suffer from a 
perception bottleneck: 
they function as static observers limited by fixed-resolution views. 
As a result, models often overlook critical relations involving small or peripheral objects and struggle to differentiate 
ambiguous predicates in complex scenes. 
To address these limitations, we propose \textbf{PSGAgent}, 
an agentic framework that enhances visual perception for PSG through a 
\textit{think--act--observe} loop. PSGAgent allows MLLMs to dynamically refine their visual perception using two tools: a \textit{Region Zoom-In Tool} for local exploration of small or boundary relations and a \textit{Visual Highlight Tool} for relation-focused segmentation that filters out irrelevant contextual noise. 
To optimize decision-making, we construct a two-stage training pipeline. First, we create a high-quality, error-aware Multimodal Chain-of-Thought (M-CoT) dataset that provides robust supervision for cold-start learning. 
Then, we design a novel hierarchical reward mechanism integrated with Group Sequence Policy Optimization (GSPO) to enable strategic and adaptive tool usage.
Extensive experiments on the PSG benchmark show that PSGAgent achieves SOTA 
performance, demonstrating strong capability in both holistic and fine-grained scene 
understanding.
\end{abstract}




\maketitle
\begin{figure}[t]
  \centering
  \includegraphics[width=\linewidth]{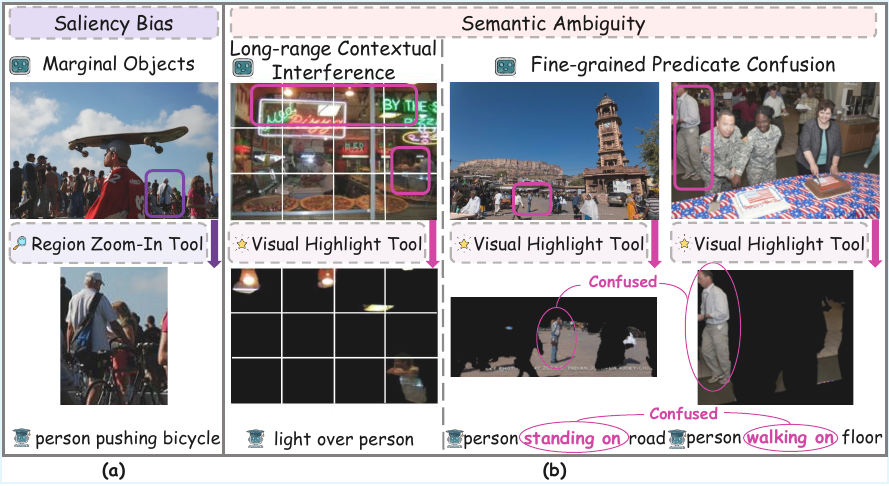} 
  \vspace{-0.6cm}
  \caption{\textbf{The perception bottleneck in PSG.} 
  \textbf{(a) Saliency bias}: static models are biased toward dominant regions, while neglecting small-scale or off-center relations. 
  \textbf{(b) Semantic ambiguity}: long-range contextual interference and fine-grained predicate confusion lead to relational uncertainty.}
  \label{fig:motivation}
  \vspace{-0.4cm}
\end{figure}
\section{Introduction}
Panoptic Scene Graph generation (PSG) serves as a critical bridge between
low-level visual perception and high-level semantic reasoning. It abstracts complex scenes into structured representations of objects
and their relations based on panoptic segmentation masks~\cite{yang2022panoptic}. 
This granular understanding is essential for downstream applications, including embodied AI~\cite{zhu2021hierarchical,gadre2022continuous,huang2025esca}, 
image-text retrieval~\cite{schroeder2020structured,liu2024semscene,wang2023scene}, and detailed image captioning~\cite{yang2023transforming,lu2025benchmarking}.

With the development of Multimodal Large Language Models 
(MLLMs)~\cite{liu2023visual,bai2025qwen2,yuan2026more,wang2026muse}, recent works have 
increasingly leveraged their powerful reasoning capabilities 
to advance scene and relation 
understanding~\cite{wang2024all}. 
For example, training-free methods~\cite{dutta2025open,li2024pixels} construct structured representations through 
multimodal prompting, but often yield suboptimal performance due to a lack of 
task-specific adaptation. 
To bridge this gap, some methods introduce visual instruction 
tuning~\cite{xu2025llavas} or RL-based architectures~\cite{chen2025compile,li2025relation,feng2026sgg} 
with rule-based rewards and textual chain-of-thought reasoning.
Despite these advances, existing methods still face 
perceptual challenges stemming from their reliance on fixed-resolution views, introducing two fundamental limitations.
First, they suffer from \textbf{saliency bias}. MLLMs tend to prioritize visually dominant regions, 
often neglecting relations involving small or peripheral objects~\cite{DBLP:conf/iclr/0002KCI25,zhang2026exploring,DBLP:journals/ijon/ZhouLZZ26}. 
For instance, the relation \textit{``person pushing bicycle''} in 
Fig.~\ref{fig:motivation}(a) 
appears near the image boundary and occupies only a small portion of the scene, 
causing the relations between the two entities to be easily overlooked.
Second, they struggle with \textbf{semantic ambiguity}. 
Accurate relation prediction is often hindered by irrelevant contextual noise, which typically manifests in two scenarios: 
(i) Long-range contextual interference: MLLMs often rely on Vision Transformers (ViTs) that serialize images into 1D sequences of patch tokens~\cite{DBLP:conf/iclr/DosovitskiyB0WZ21}. 
When two interacting entities are spatially distant (e.g., \textit{``light''} and \textit{``person''}), 
their corresponding tokens are separated by a vast number of intervening background patches. 
Under a fixed global view, this dense visual noise dominates the self-attention mechanism, 
severely obscuring the cross-patch relational cues necessary to bridge the spatial gap. 
(ii) Fine-grained predicate confusion: subtle actions such as \textit{``standing on''} versus \textit{``walking on''} often exhibit highly similar global appearances, making them difficult to distinguish without examining local visual details.
Current MLLM-based frameworks~\cite{wang2024all,dutta2025open,li2024pixels,xu2025llavas,chen2025compile,li2025relation,feng2026sgg} process visual inputs at a fixed, global granularity.
\textbf{They inherently lack the mechanism to adaptively re-examine localized visual details, which leaves both saliency bias and semantic ambiguities unresolved}.

To address these limitations, we propose \textbf{PSGAgent}, an agentic MLLM framework for 
PSG. 
Instead of focusing on fixed-resolution views, PSGAgent follows a \textit{think--act--observe} loop: the model generates reasoning thoughts and selectively invokes perception tools to obtain visual observations. 
Specifically, PSGAgent employs two complementary tools: a \textit{Region Zoom-In Tool} to mitigate 
saliency bias, which helps recover relations involving small or peripheral objects, and a 
\textit{Visual Highlight Tool} to reduce semantic ambiguity by suppressing long-range contextual interference and emphasizing interaction regions.
Through iterative interactions with these tools, PSGAgent progressively refines its understanding of the scene and produces more accurate relational predictions.
Training such an agent raises a key challenge: 
learning \emph{when} and \emph{how} to invoke tools effectively. Standard RL is hindered by sparse rewards and unstable training, while large-scale trajectory annotation is prohibitively costly.
We address this with a two-stage training strategy. 
First, we construct error-aware Multimodal Chain-of-Thought (M-CoT) trajectories from hard PSG samples, which provide structured supervision for tool-use behavior during SFT. 
Second, we employ Group Sequence Policy Optimization (GSPO)~\cite{zheng2025group} with hierarchical 
rewards to stabilize RL training. 
Our contributions are as follows:
\begin{itemize}
    \item We identify the perception bottleneck (saliency bias and semantic ambiguity) in existing methods, and alleviate it by reformulating PSG as a multi-turn reasoning process.
    \item We propose \textbf{PSGAgent}, an agentic framework featuring task-specific tools and a hierarchical reward mechanism optimized via RL to enable adaptive agentic reasoning.
    \item We introduce a high-quality error-aware M-CoT dataset tailored for agent training, which enables efficient cold-start learning and improves training stability.
    \item PSGAgent achieves SOTA performance on the PSG benchmark, demonstrating improved capabilities in holistic and fine-grained relation understanding.
\end{itemize}

\section{Related Work}

\subsection{Panoptic Scene Graph Generation}

PSG represents complex scenes as structured object--relation graphs based on panoptic segmentation
masks~\cite{yang2022panoptic}. 
 Traditional research includes two-stage pipelines like MOTIFS~\cite{zellers2018neural} and VCTree~\cite{tang2019learning} 
 or one-stage frameworks such as PSGTR and 
  PSGFormer~\cite{yang2022panoptic}. 
  Building upon these methods, CAFE~\cite{shi2025easy} incorporates shape-aware features via 
  curriculum learning. ADTrans~\cite{li2024panoptic} utilizes semantics-prototype learning 
  to transfer biased annotations into informative targets. 
Nevertheless, the generalization capacity of these models is severely constrained by the limited diversity of manual annotations.

Recognizing the constraints of closed-set vocabularies, 
recent research leverages MLLMs 
for open-vocabulary relation comprehension. OwSGG~\cite{dutta2025open}
 proposes a training-free framework relying on multimodal prompting, 
 whereas OvSGTR~\cite{chen2024expanding} employs 
 transformer decoders to predict open-set relations. 
 With recent advances in RL~\cite{tan2026rado,chu2026towards}, 
 frameworks increasingly combine SFT with RL to enhance reasoning. Relation-R1~\cite{li2025relation} utilizes cognitive CoT 
 guidance, SGG-R$^3$~\cite{feng2026sgg} adopts a three-stage structured reasoning 
 pipeline and R1-SGG~\cite{chen2025compile} applies rule-based 
 rewards to enforce structural schema consistency. 
 However, both traditional and MLLM-based methods largely rely on fixed views, which limits their ability to address the perception bottleneck targeted in this work.
 \subsection{Thinking with Images in MLLMs}

Recent MLLMs increasingly integrate visual evidence into reasoning, 
a paradigm termed \textit{Thinking with Images}~\cite{DBLP:conf/eccv/WuWTWHOTW24,DBLP:conf/nips/WuMZ000W24,DBLP:conf/eccv/LiuCLZLRZYSZZGL24,DBLP:conf/cvpr/WuX24a,DBLP:conf/nips/ZhengLCZLLLLZ24,DBLP:conf/emnlp/Cheng0W024,DBLP:conf/iclr/ZengAICWWTPRSLV23,DBLP:conf/iccv/Shtedritski0V23,DBLP:journals/corr/abs-2505-23558}. 
Specifically, DeepEyes~\cite{zheng2026deepeyes} and Pixel Reasoner~\cite{su2025pixel} employ intrinsic 
rewards to incentivize active perception operations. 
VisionThink~\cite{yang2025visionthink} and VisionReasoner~\cite{liu2026visionreasoner} leverage RL to optimize visual resolution and unified perception tasks, respectively. 
For more complex scenarios, WebWatcher~\cite{geng2026webwatcher} interleaves text with visual steps to 
solve deep-research problems and 
VTool-R1~\cite{wu2026vtoolr} uses Python-based tool-calling to enhance multimodal reasoning.

However, directly applying these agentic frameworks to PSG remains difficult, as existing methods struggle to simultaneously address two critical challenges. 
First, existing methods~\cite{su2025pixel,geng2026webwatcher,wu2026vtoolr} are primarily query-driven, 
where visual exploration is guided by specific prompts. 
In contrast, PSG is a proactive task that requires 
exhaustive discovery of valid relations across the scene, including dense stuff--thing interactions and long-tail predicates, without explicit query supervision. 
Second, prior tool-use approaches~\cite{fan2025grit,zheng2026deepeyes,liu2026visionreasoner} mainly focus on individual entities 
(e.g., region cropping for object recognition) or coarse-grained predicates (e.g., left, right), 
whereas PSG demands reasoning over the fine-grained interaction between objects to resolve semantic ambiguity. 
\begin{figure*}
  \centering
  \includegraphics[width=\linewidth]{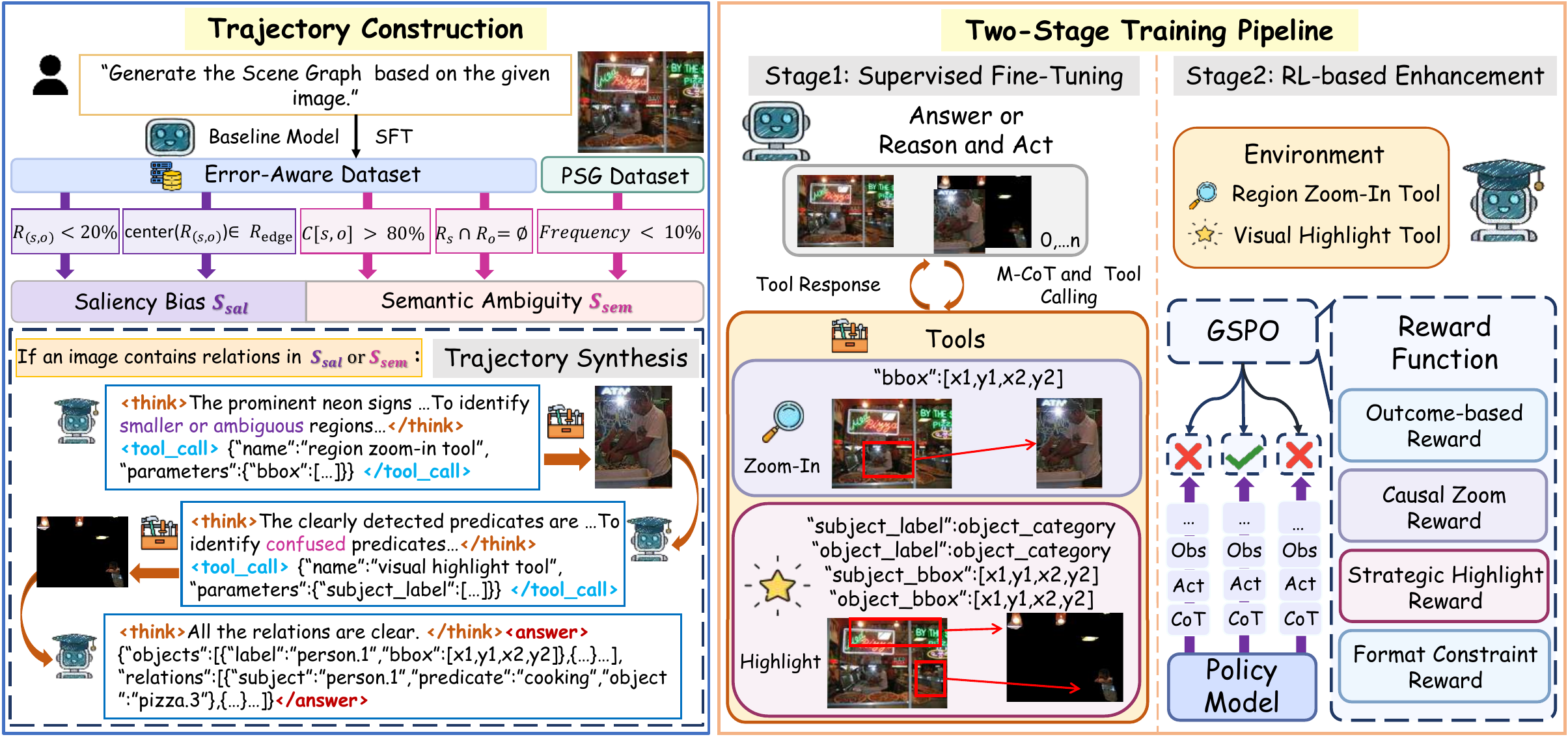} 
  \caption{\textbf{Overview of PSGAgent.} We reformulate PSG as 
  a multi-turn, tool-augmented decision-making process.
  \textbf{(Left) Trajectory Construction.} 
  We perform error-aware data mining to identify challenging cases caused by saliency bias 
  (relations occupying small regions or located near image boundaries) and semantic ambiguity 
  (long-range contextual interference, high predicate confusion, or tail relations). These identified cases are used to synthesize error-aware Multimodal 
  Chain-of-Thought (M-CoT) trajectories through a structured generation pipeline. \textbf{(Right) Cold-start SFT and RL-based Enhancement.} 
The agent is first initialized via SFT to learn basic tool usage, and then 
optimized with GSPO for strategic reasoning. Through a \textit{think--act--observe} interaction loop guided by a hybrid reward 
function ($R_{\text{outcome}}$, $R_{\text{causal}}$, $R_{\text{highlight}}$, and $R_{\text{format}}$), the agent learns to autonomously invoke tools (\textit{Region Zoom-In Tool} and \textit{Visual Highlight Tool}) to achieve holistic and fine-grained scene understanding.}
  \label{fig:framework}
\end{figure*}
\section{Method}

We introduce PSGAgent, formulating PSG as a multi-turn agentic reasoning process (Fig.~\ref{fig:framework}). 
Given an image $\mathcal{I} \in \mathbb{R}^{H \times W \times 3}$, the agent models the distribution $P(\mathcal{G}|\mathcal{I})$ to generate a panoptic scene graph $\mathcal{G} = (\mathcal{O}, \mathcal{R})$. 
The object set $\mathcal{O} = \{o_i = (c_i, m_i)\}_{i=1}^N$ encapsulates both thing and stuff entities, each grounded by a category $c_i$ and a non-overlapping binary mask $m_i \in \{0, 1\}^{H \times W}$. 
The relation set $\mathcal{R} = \{r_{i,j}\}$ captures the set of pairwise relationships, denoting the predicate category $r_{i,j}$ between the $i$-th subject and $j$-th object.

Our framework features three core components: custom perception tools 
(\textit{Region Zoom-In Tool} and \textit{Visual Highlight Tool}) (Sec.~\ref{sec:AgenticActivePerceptionTools}), error-aware SFT (Sec.~\ref{sec:TrajectoryColdStartSFT}) and GSPO with hierarchical rewards (Sec.~\ref{sec:AgenticReinforcementLearning}). 
Through this design, the agent dynamically invokes
tools to resolve ambiguities, and iteratively generates relational
triplets until the full scene graph is completed.

\subsection{Agentic Active Perception Tools}
\label{sec:AgenticActivePerceptionTools}
We equip the agent with two tools that modify the observation to resolve perceptual uncertainty. 
Each tool is formulated as a visual operator:
$
\mathcal{I}' = \mathcal{T}(\mathcal{I}; \phi),
$
where $\phi$ denotes tool-specific parameters.

\subsubsection{Region Zoom-In Tool.}
Given a bounding box $b$, the tool crops the specified region to mitigate saliency bias:
\begin{equation}
\mathcal{I}_{\text{zoom}} = \mathcal{T}_{\text{zoom}}(\mathcal{I}; b) = \text{Crop}(\mathcal{I}, b).
\end{equation}

\subsubsection{Visual Highlight Tool.}
Given subject and object masks $\mathcal{M}_s$ and $\mathcal{M}_o$, the highlight tool suppresses irrelevant regions:
\begin{equation}
\mathcal{I}_{\text{highlight}} =
\mathcal{T}_{\text{highlight}}(\mathcal{I}; s,o)
= \mathcal{I} \odot (\mathcal{M}_s \cup \mathcal{M}_o),
\end{equation}
where $\odot$ denotes the Hadamard product and $\cup$ denotes 
the union of subject and object masks. 
Unlike entity-centric cropping ~\cite{fan2025grit,zheng2026deepeyes,liu2026visionreasoner}, 
our \textit{Visual Highlight Tool} suppresses task-irrelevant visual noise,
allowing the agent to focus purely on inter-object interactions, effectively resolving semantic ambiguity in complex scenes.

\subsection{Trajectory Construction and Cold-Start Supervised Fine-tuning}
\label{sec:TrajectoryColdStartSFT}
Pure RL struggles with sparse rewards for complex tool-use behaviors. 
To provide a stable initialization, we construct an M-CoT dataset $\mathcal{D}_{\text{SFT}}$ through error-aware data mining, focusing on samples where fixed perception typically fails.

\subsubsection{Error-Aware Data Mining.}

We first train a vanilla MLLM (e.g., Qwen2.5-VL-3B~\cite{bai2025qwen2} or LLaVA~v1.5~\cite{liu2024improved}) on the PSG training set and collect mispredicted triplets to form an error-aware dataset. 
From this set, we further identify two types of challenging cases that 
can be addressed 
by our perception tools:

\begin{itemize}
    \item \textbf{Saliency Bias ($\mathcal{S}_{\text{sal}}$).}
    Triplets are considered saliency-biased if the 
    union region of the subject and object bounding boxes, $\mathcal{R}_{(s,o)}$, 
    occupies less than $20\%$ 
    of the image area or the center of the union region, $\mathrm{center}(\mathcal{R}_{(s,o)})$, 
    lies near the image boundary. 
    These small-scale or edge-located relations are prone to be 
    overlooked by fixed perception and therefore trigger the \textit{Region Zoom-In Tool}.

    \item \textbf{Semantic Ambiguity ($\mathcal{S}_{\text{sem}}$).}
    We first select triplets whose subject and object bounding boxes do not overlap as candidates. 
    Predicates are then considered ambiguous if they (i) exhibit high confusion in the normalized confusion matrix $C$ (i.e., $C[i,j]>\mu$ for some $j\neq i$) where $C[i,j]$ denotes the proportion of samples with ground-truth relation $i$ but predicted as $j$, or (ii)  belong to the low-frequency
    tail of the predicate distribution. 
    Specifically, we categorize
    predicates into head and tail groups
    based on their training frequency to analyze long-tail
    relation recognition.
    Triplets satisfying these conditions are assigned to $\mathcal{S}_{\text{sem}}$ and handled by the \textit{Visual Highlight Tool} to isolate the interacting pair from cluttered backgrounds.
\end{itemize}

\subsubsection{Trajectory Synthesis.}

As illustrated in the bottom-left of 
Fig.~\ref{fig:framework}, we synthesize a multi-turn M-CoT trajectory for each 
identified hard sample to simulate a dynamic perception process. Instead of producing a direct prediction, 
the trajectory follows a \textit{think--act--observe} paradigm similar to ReAct~\cite{yao2022react}. 

At each step $i$, the assistant follows a structured interaction loop consisting of three components: 
(i) a thought, the agent's intermediate reasoning used to analyze perceptual limitations (e.g., saliency bias or semantic ambiguity), enclosed in \texttt{<think>} \dots \texttt{</think>}; 
(ii) an action, where the agent invokes a perception tool (e.g., \textit{Region Zoom-In Tool} or \textit{Visual Highlight Tool}) wrapped in \texttt{<tool\_call>} \dots \texttt{</tool\_call>}, 
or produces the final prediction; 
and (iii) an observation, the environment feedback containing the enhanced visual input, 
returned within \texttt{<tool\_response>} \dots \texttt{</tool\_response>}. 
This \textit{think--act--observe} process iterates until the agent outputs the scene graph $\mathcal{G}$ or 
reaches the maximum number of iterations
$u_{\mathrm{iter}}$.

Formally, a trajectory of length $L$ is represented as
\begin{equation}
\tau = \{(t_1, a_1, o_1), (t_2, a_2, o_2), \dots, (t_L, a_L, o_L)\}\quad L \le u_{\mathrm{iter}},
\end{equation}
where $t_i$ denotes the thought at step $i$, $a_i$ denotes the action and $o_i$ denotes the returned observation. 
By fine-tuning 
MLLMs (e.g., Qwen2.5-VL-3B~\cite{bai2025qwen2}) on these trajectories, the agent learns to 
navigate complex scenes and execute strategic tool 
calls.

\subsection{Agentic Reinforcement Learning}
\label{sec:AgenticReinforcementLearning}
\subsubsection{Agentic RL with GSPO}
We employ GSPO ~\cite{zheng2025group}, which
mitigates the high-variance noise inherent in long-form generation by shifting optimization to the sequence level. 
Given an input state $x$ (comprising the image $\mathcal{I}$ and interaction history), we sample $G$ reasoning trajectories $\{y_i\}_{i=1}^G \sim \pi_{\theta_{old}}(\cdot|x)$ and optimize:
\begin{equation}
\begin{split}
J_{GSPO}(\theta) =& \mathbb{E}_{x, \{y_i\} \sim \pi_{\theta_{old}}} \Bigg[ \frac{1}{G} \sum_{i=1}^{G} \min \Big( s_i(\theta) A_i, \\
&\text{clip}(s_i(\theta), 1-\epsilon, 1+\epsilon) A_i \Big) \Bigg],
\end{split}
\label{eq:gspo_obj}
\end{equation}
where
$A_i=\frac{r_i-\operatorname{mean}(\{r_i\})}
{\operatorname{std}(\{r_i\})+\sigma}$
is the group-normalized advantage and $r_i=r(x,y_i)$.
The sequence-level importance ratio is the geometric mean of token-level policy ratios:
\begin{equation}
s_i(\theta) = \exp \left( \frac{1}{|y_i|} \sum_{t=1}^{|y_i|} \log \frac{\pi_\theta(y_{i,t} | x, y_{i,<t})}{\pi_{\theta_{old}}(y_{i,t} | x, y_{i,<t})} \right).
\label{eq:seq_ratio}
\end{equation}

\subsubsection{Outcome-Based Reward ($R_{\text{outcome}}$).}

We define an outcome-based reward that jointly encourages 
accuracy and tool usage:

\begin{equation}
R_{\text{outcome}} =
\mathbb{I}_{\text{valid}}
\left[
4\mathcal{A}
+
\mathbb{I}_{N_{\text{img}} > 1}\,(0.5 + 1.5\mathcal{A})
\right],
\end{equation}
where $\mathbb{I}_{\text{valid}}$ indicates JSON format validity and $\mathbb{I}_{N_{\text{img}} > 1}$ denotes successful tool invocation. 
The triplet accuracy is defined as

\begin{equation}
\mathcal{A} =
\frac{1}{|\mathcal{G}_{\text{GT}}|}
\sum_{(s_i,p_i,o_i) \in \hat{\mathcal{G}}_p}
\mathbb{I}
\left[
(s_i,p_i,o_i)\approx\mathcal{G}_{\mathrm{GT}}
\right],
\end{equation}
where $\mathcal{G}_{\text{GT}}$ denotes the ground-truth triplet set and
$\hat{\mathcal{G}}_p$ the predicted  set.
A match $(s_i,p_i,o_i) \approx \mathcal{G}_{\text{GT}}$ requires
(i) exact category alignment of subject, predicate, and object, and
(ii) IoU $\ge 0.5$ between the predicted entities and the ground-truth bounding boxes.

\subsubsection{Causal Zoom Reward ($R_{\text{causal}}$)}

We assign a reward only when the zoomed region actively helps recover valid relations. For a zoom action at step $t$ producing region $\mathcal{R}_t$, the reward is defined as:
\begin{equation}
\small
\begin{aligned}
r_t^{\text{zoom}} 
=&\;
\underbrace{\mathbb{I}\!\left[
\exists (s_i,p_i,o_i) \in \hat{\mathcal{G}}_p \cap \mathcal{G}_{\text{GT}}
\;\text{s.t.}\; \mathcal{R}_{(s_i,o_i)} \subset \mathcal{R}_t
\right]}_{\text{relation recovery}}
\cdot
\underbrace{\mathbb{I}\!\left[
\frac{|\mathcal{R}_t|}{|\mathcal{I}|} < \tau_{\text{ratio}}
\right]}_{\text{scale constraint}} \\
&+
\underbrace{\left(1 + \lambda_{\text{edge}}
\mathbb{I}[\mathrm{center}(\mathcal{R}_t) \in \mathcal{R}_{\text{edge}}]\right)}_{\text{boundary bonus}}\\
&-
\underbrace{\lambda_{\text{red}}
\max_{k<t}
\mathbb{I}\!\left[
\mathrm{IoU}(\mathcal{R}_t, \mathcal{R}_k) > \tau_{\text{iou}}
\right]}_{\text{redundancy penalty}}.
\end{aligned}
\end{equation}
Here, $\mathcal{R}_{(s_i,o_i)}$ denotes the union of the subject and object bounding boxes. The first two terms reward zoom actions only if the cropped region is sufficiently small (bounded by $\tau_{\text{ratio}}$) and successfully encapsulates at least one correctly predicted ground-truth triplet. The boundary bonus encourages the exploration of peripheral areas ($\mathcal{R}_{\text{edge}}$), while the redundancy penalty discourages repeated observations ($\mathrm{IoU} > \tau_{\text{iou}}$). 
Unlike traditional query-driven 
agents ~\cite{su2025pixel,geng2026webwatcher,wu2026vtoolr} that terminate upon finding a single localized target, 
the combination of boundary bonus and redundancy penalties encourages 
our agent to adopt a \emph{proactive} exploration paradigm, 
promoting a more comprehensive exploration of the scene.

\subsubsection{Strategic Highlight Reward ($R_{\text{highlight}}$)}
When the agent invokes the \textit{Visual Highlight Tool} for a subject-object pair $(s_i,o_i)$ at time step $t$, the reward is defined as:
\begin{equation}
\small
\begin{aligned}
r_t^{\text{highlight}}
=
\underbrace{
\mathbb{I}\!\left[
(s_i,p_i,o_i) \in \hat{\mathcal{G}}_p \cap \mathcal{G}_{\text{GT}}
\right]
}_{\text{triplet correctness}} + 
\underbrace{
\left(1 + \lambda_{\text{diff}}
\mathbb{I}[p_i \in \mathcal{S}_{\text{sem}}]
\right)
}_{\text{difficulty bonus}} .
\end{aligned}
\end{equation}
The first term rewards the agent when the highlighted triplet is both predicted by the agent and accurately matches a ground-truth relation. The second term assigns an additional bonus when the predicted predicate $p_i$ belongs to the semantically ambiguous set $\mathcal{S}_{\text{sem}}$. Overall, this reward design encourages
the agent to strategically employ the highlight operation to disambiguate challenging predicates and alleviate the long-tail problem.

\subsubsection{Format Constraint Reward ($R_{\text{format}}$).}
We penalize outputs that violate syntactic or 
grounded constraints, including: (i) syntax errors, such as 
invalid JSON or mismatched M-CoT tags 
(\texttt{<think>}, \texttt{<tool\_call>}); and (ii) visual 
hallucinations, where the agent predicts object identifiers or 
categories not present in the visual observation.
The total reward is a weighted sum: \begin{equation}
    R_{\text{total}} = R_{\text{outcome}} + \alpha_{\text{causal}} R_{\text{causal}} + \alpha_{\text{highlight}} R_{\text{highlight}} + \gamma_{\text{format}}R_{\text{format}},
    \end{equation} where $\alpha_{\text{causal}}$, $\alpha_{\text{highlight}}$, and $\gamma_{\text{format}}$ control the trade-off between perception strategy and output stability.

\section{Experiments}

\subsection{Experimental Settings}

\subsubsection{Datasets.} We evaluate our model on the 
PSG dataset~\cite{yang2022panoptic}, 
which contains 48,749 images with 133 object categories and 
56 predicate categories. It provides high-quality panoptic segmentation masks for grounded relationship triplets.

\subsubsection{Tasks.} Following standard scene graph 
generation protocols~\cite{yang2022panoptic}, we evaluate our model on two 
primary tasks: (i) Predicate Classification (PredCls): 
Given all ground-truth object boxes and labels, 
the model predicts the pairwise predicate categories. 
This task isolates the model's relational reasoning ability 
from its perception accuracy. 
(ii) Scene Graph Detection (SGDet): Given an input image, the model detects all objects (predicting both bounding boxes and categories) and infers their pairwise predicates from scratch.

\subsubsection{Metrics.} To evaluate performance, 
we adopt several standard metrics. 
We report Recall@100 (Recall), mean Recall@100 (mRecall) and their average (Mean).
Recall@100 measures the proportion of ground-truth triplets retrieved among the top-100 predictions, while mean Recall@100 averages class-wise recall across predicate categories to reduce bias toward frequent relations.
We further report AP@50 for entity localization and classification quality, and zero-shot Recall@100 (zs-Recall)
to assess generalization to unseen triplets. 
Following prior work~\cite{chen2024expanding,li2024pixels,he2022towards}, we evaluate Open-Vocabulary Relation (OvR) prediction by splitting object and relation categories into base and novel groups at a 7:3 ratio.
\subsubsection{Implementation Details.}  We use Qwen2.5-VL-3B-Instruct~\cite{bai2025qwen2} as the backbone
and obtain subject and object masks for the \textit{Visual Highlight Tool} using a pretrained Mask2Former~\cite{cheng2022masked}.
Training consists of cold-start SFT followed by RL.
During SFT, 
we train on the constructed 8K M-CoT dataset for 5 
epochs with a batch size of 32 using the AdamW optimizer with a 
learning rate of $1 \times 10^{-5}$. For RL, GSPO samples $G=3$ trajectories per query.
We set $\tau_{\text{ratio}}=0.2$, $\tau_{\text{iou}}=0.8$, $\lambda_{\text{edge}}=\lambda_{\text{red}}=\lambda_{\text{diff}}=0.5$, $\alpha_{\text{causal}}=\alpha_{\text{highlight}}=0.8$, and $\gamma_{\text{format}}=0.5$.
During inference, the agent 
performs up to $u_{\mathrm{iter}}=5$ \textit{think--act--observe} interaction steps, with temperature set to 0.7. 
All experiments are conducted on four A6000 GPUs, and GSPO training converges in approximately 7 days.

\begin{table}
  \caption{Comparison with state-of-the-art methods on the 
  PSG benchmark under the SGDet setting. * denotes using predefined object classes and relation categories in prompts. Zm. T. and Hl. T. denote \textit{Region Zoom-In Tool} and \textit{Visual Highlight Tool}, respectively. Bold and underline results highlight the best and second-best.}
  \label{tab:sgg_main}
  \vspace{-0.1cm}
  \centering
  \small
  \footnotesize 
  \setlength{\tabcolsep}{2.8pt}
  \begin{tabular}{l c c c c c} 
    \toprule
    \multirow{2}{*}{\textbf{Model}} & \multirow{2}{*}{\textbf{Params}}&\multicolumn{4}{c}{\textbf{SGDet}} \\
    \cmidrule(lr){3-6}
     &&\textbf{AP@50$\uparrow$} & \textbf{Recall$\uparrow$} & \textbf{mRecall$\uparrow$} & \textbf{Mean$\uparrow$} \\
    \midrule
    \rowcolor{gray!10} 
    \multicolumn{6}{l}{\textit{Specialized PSG Models}} \\
    
    IMP~\cite{xu2017scene}            & -- & -- &18.60&7.23&12.92 \\
    MOTIFS~\cite{zellers2018neural}      & -- & -- &22.00&9.69&15.85\\
    VCTree~\cite{tang2019learning}      & -- & -- &22.50&10.20&16.35\\
    GPSNet~\cite{lin2020gps}      & -- & -- &20.10&7.69&13.90\\
    PSGFormer~\cite{yang2022panoptic}& -- & -- &20.20&18.70&19.45\\
    HiLo\cite{zhou2023hilo}               & -- & -- &43.00&33.10&38.05\\
    PairNet\cite{wang2024pair}            & -- & -- &39.60&30.60&35.10\\
    DSGG\cite{hayder2024dsgg}               & -- & -- &50.00&\textbf{44.40}&\underline{47.20}\\
    \midrule
    \rowcolor{gray!10} 
    \multicolumn{6}{l}{\textit{MLLM-based Methods}} \\
    TextPSG~\cite{zhao2023textpsg}               & --  & --    & 4.82  & --    & --\\
    LLaVA~v1.5~\cite{liu2024improved}              & 7B  & 0.07  & 0.00  & 0.00  & 0.00\\
    ASMv2~\cite{wang2024all}                   & 13B & 21.45 & 14.77 & 11.82 & 13.30\\
    OvSGTR~\cite{chen2024expanding}               & --  & --    & 41.40 & 28.30 & 34.90\\
    PGSG~\cite{li2024pixels}             & 13B & --    & 33.40 & 22.10 & 27.75\\
    LLaVA-SpaceSGG~\cite{xu2025llavas}       & 13B & --    & 15.43 & 13.23 & 14.33\\
    OwSGG~\cite{dutta2025open}           & 72B & --    & 13.98 & 13.67 & 13.83\\
    R1-SGG*~\cite{chen2025compile}       & 2B  & 39.28 & 38.49 & 31.21 & 34.85\\
    R1-SGG*~\cite{chen2025compile}       & 7B  & 42.05 & 43.48 & 33.71 & 38.60\\
    Relation-R1~\cite{li2025relation}    & 3B  & 25.87 & 21.32 & 23.60 & 22.46\\
    SGG-R$^3$~\cite{feng2026sgg}         &3B   &38.82& 47.63 & 38.33 & 42.98\\
    Qwen2.5-VL-3B-Instruct               & 3B  & 5.12  & 0.57  & 0.31  & 0.44 \\
    \quad + SFT                          & 3B  & 30.10 & 21.67 & 10.37 & 14.76\\
    Qwen2.5-VL-72B-Instruct*             & 72B & 28.32 & 12.37 & 9.91  & 11.14\\
    \quad + ZM.T. HL.T.*                   & 72B & 27.57 & 14.48 & 14.17  & 14.33\\
    \midrule
    \rowcolor{gray!10} 
    \multicolumn{6}{l}{\textit{Thinking With Images Methods}} \\
    DeepEyes~\cite{zheng2026deepeyes}    & 3B  & 44.52 & 43.18 & 30.10 & 36.64\\
    Pixel Reasoner~\cite{su2025pixel}    & 3B  & 44.19 & 43.73 & 29.67 & 36.70\\
    VisionReasoner ~\cite{liu2026visionreasoner} & 3B  & 40.13 & 38.27 & 24.22 & 31.25\\
    VisionThink~\cite{yang2025visionthink}   & 3B  & 35.63 & 35.64 & 20.02 & 27.83\\
    \midrule
    
    \rowcolor{gray!10} 
    PSGAgent (Ours)         & 3B & \underline{49.91} & \underline{50.27} & 42.88&46.58\\
    \rowcolor{gray!10}
    \textbf{PSGAgent* (Ours)} & 3B & \textbf{50.41} & \textbf{51.81} & \underline{44.19} & \textbf{48.00}\\
    \bottomrule
    \vspace{-0.4cm}
  \end{tabular}
  
\end{table}

\subsection{Main Results}
\subsubsection{Comparison under SGDet.}
As Table~\ref{tab:sgg_main} shows, our method achieves SOTA among all MLLM-based frameworks. 
Specifically, the standard PSGAgent(3B) obtains 50.27\% Recall, 42.88\% mRecall and 46.58\% Mean, 
effectively alleviating the perception bottleneck in MLLMs. 
When leveraging predefined category prompts, \textbf{PSGAgent*} further pushes the performance to 51.81\% Recall, 44.19\% mRecall and 48.00\% Mean. 
Notably, PSGAgent* successfully surpasses the strongest specialized model, DSGG~\cite{hayder2024dsgg} (Mean 47.20\%).
While DSGG's explicit long-tail optimizations for closed-set distributions yield a marginal edge in mRecall (44.40\% vs. 44.19\%), 
these slight gains come at the expense of broader applicability. 
In contrast, PSGAgent achieves higher Recall and Mean scores,
demonstrating that an agentic MLLM framework can surpass architectures heavily engineered for 
specific, narrow scenarios.

Compared to recent MLLMs, PSGAgent demonstrates clear advantages. While standard reasoning models like R1-SGG~\cite{chen2025compile} require scaling up to 7B parameters to achieve a 
38.60\% Mean, our 3B PSGAgent outperforms it by a significant margin (+7.98\% Mean). Furthermore, when compared to the baseline model
(Qwen2.5-VL-3B-Instruct + SFT, Mean 14.76\%), our agentic framework yields a substantial performance improvement.
 This highlights that simply fine-tuning an MLLM is insufficient, 
 whereas our \textit{think--act--observe} loop effectively captures both dominant and long-tail relations.

To validate our tools' effectiveness, we evaluate the Qwen2.5-VL-72B-Instruct in a zero-shot setting. 
Despite its scale, the vanilla 72B model achieves an mRecall of 9.91\%, 
showing that parameter scaling alone cannot fully address the perception bottleneck. 
However, simply equipping the frozen 72B model with our \textit{Region Zoom-In Tool} and \textit{Visual Highlight Tool} 
improves the mRecall to 14.17\%. 
This confirms the plug-and-play utility of our dynamic refinement tools. 
More crucially, our 3B PSGAgent* (Mean 48.00\%) significantly outperforms the tool-augmented 72B model, 
suggesting that strategic tool-use learning via our agentic framework is more critical for PSG than raw model scaling.

We also evaluate \emph{thinking with images} methods. Since these methods~\cite{zheng2026deepeyes,su2025pixel,liu2026visionreasoner,yang2025visionthink} were not originally designed for PSG,
we retrain them on the PSG dataset for cold-start SFT and use the same outcome-based reward for RL. 
Although Pixel Reasoner~\citep{su2025pixel} performs best among these adapted baselines, PSGAgent still improves the Mean score from 36.70\% to 46.58\%.
This gap suggests that query-driven or object-centric agentic frameworks are insufficient for exhaustive structured relation discovery, 
while our PSGAgent better supports holistic PSG reasoning.
\begin{table}
  \caption{Comparison with state-of-the-art methods on the PSG benchmark under the PredCls setting.}
  \label{tab:predcls_main}
  \centering
  \small 
  \footnotesize
  \setlength{\tabcolsep}{2.8pt}
  \vspace{-0.1cm}
  \begin{tabular}{l c c c c} 
    \toprule
    \multirow{2}{*}{\textbf{Model}} & \multirow{2}{*}{\textbf{Params}} & \multicolumn{3}{c}{\textbf{PredCls}} \\
    \cmidrule(lr){3-5}
     & & \textbf{Recall$\uparrow$} & \textbf{mRecall$\uparrow$} & \textbf{Mean$\uparrow$} \\
    \midrule
    \rowcolor{gray!10} 
    \multicolumn{5}{l}{\textit{Specialized PSG Models}} \\
    IMP~\cite{xu2017scene}            & -- &38.90&11.60&25.25\\
    GPSNet~\cite{lin2020gps}      & -- &44.70&18.30&31.50\\
    C-SCC~\cite{jin2023fast}              & -- &46.50&36.40&41.45\\
    MOTIFS~\cite{zellers2018neural}      & -- &52.40&22.90&37.65\\
    MOTIFS + CAFE~\cite{shi2025easy}     & -- &47.20&36.90&42.05\\
    VCTree~\cite{tang2019learning}      & -- &\underline{52.70}&23.30&38.00\\
    VCTree + CAFE~\cite{shi2025easy}     & -- &48.10&\underline{37.60}&\underline{42.85}\\
    \midrule
    \rowcolor{gray!10} 
    \multicolumn{5}{l}{\textit{MLLM-based Methods}} \\
    TextPSG~\cite{zhao2023textpsg}            & -- & -- &12.24 & -- \\
    \midrule
    \rowcolor{gray!10} 
    \textbf{PSGAgent (Ours)}            & 3B &\textbf{73.04} & \textbf{67.75}&\textbf{70.40}\\
    \bottomrule
  \end{tabular}
\end{table}

\subsubsection{Results under PredCls.}
Table~\ref{tab:predcls_main} presents the evaluation under the PredCls setting, which isolates relation prediction by providing ground-truth object regions. 
In this idealized setting, PSGAgent achieves 73.04\% Recall and 67.75\% mRecall, 
establishing a substantial lead over all previous specialized PSG models. 
Specifically, PSGAgent improves the Mean performance from 42.85\% (VCTree + CAFE) to 70.40\%, an absolute gain of 27.55\%.
This result indicates that PSGAgent is particularly effective at relation reasoning once reliable object regions are available. 
The strong mRecall also suggests that our framework better handles long-tail predicate distributions, which remain challenging in scene graph generation.
\begin{table}
  \caption{Results of zs-Recall and Open-Set (OvR) on PSG under the SGDet setting.}
  \vspace{-0.1cm}
  \label{tab:zr_ngr}
  \centering
  \small
  \footnotesize 
  \setlength{\tabcolsep}{2.8pt}
  \begin{tabular}{l c c ccc}
    \toprule
    \multirow{2}{*}{\textbf{Model}} & \multirow{2}{*}{\textbf{Params}} &\textbf{zs-} & \multicolumn{3}{c}{\textbf{Open-Set (OvR)}}\\
    \cmidrule(lr){4-6}
    & &\textbf{Recall}$\uparrow$&\textbf{Recall}$\uparrow$&\textbf{mRecall}$\uparrow$&\textbf{Mean}$\uparrow$\\
    \midrule
    \rowcolor{gray!10} 
    \multicolumn{6}{l}{\textit{Specialized PSG Models}} \\
    PSGFormer~\cite{yang2022panoptic}        &--&4.90&3.10&2.70&2.90     \\
    PSGTR~\cite{yang2022panoptic}            &--&6.40&3.30&2.00&2.65     \\
    SGTR~\cite{li2022sgtr}             &--&5.80& -- & -- & --      \\
    PairNet~\cite{wang2024pair}          &--& -- &6.50&5.10&5.80      \\
    HiLo~\cite{zhou2023hilo}             &--& -- &4.80&3.60&4.20      \\
    DSGG~\cite{hayder2024dsgg}             &--& -- &5.60&5.00&5.30      \\
    \midrule
    \rowcolor{gray!10} 
    \multicolumn{6}{l}{\textit{MLLM-based Methods}} \\
    OvSGTR~\cite{chen2024expanding}  &--& --  &\underline{22.80}&\underline{14.00}&\underline{18.40}\\
    PGSG~\cite{li2024pixels}             &13B&\textbf{8.90}&17.70&11.50&14.60        \\
    OwSGG~\cite{dutta2025open}            &72B&6.44&13.54&13.35&13.45\\
    R1-SGG~\cite{chen2025compile}           &7B &7.70& --  & --  & --           \\
    SGG-R$^3$~\cite{feng2026sgg}        &3B &7.70& --  & --  & --           \\
    \midrule
    \rowcolor{gray!10} 
    \textbf{PSGAgent (Ours)}                &3B &\underline{8.19}&\textbf{27.52}&\textbf{20.20}&\textbf{23.86} \\
    \bottomrule
  \end{tabular}
\end{table}

\subsubsection{Zero-shot and Open-set Evaluation.}
To evaluate generalization ability, Table~\ref{tab:zr_ngr} reports zero-shot Recall and open-set performance under the OvR setting. 
PSGAgent achieves a competitive zs-Recall of 8.19\%, significantly outperforming specialized PSG models and rivaling the PGSG~\cite{li2024pixels} despite its 13B parameter size. 
This underscores the model's superior parameter efficiency and zero-shot generalization capabilities.
Regarding the more challenging open-set evaluation, PSGAgent achieves the best results across all OvR metrics, with 27.52\% Recall, 20.20\% mRecall, and 23.86\% Mean, 
significantly outperforming prior MLLM-based methods such as PGSG~\cite{li2024pixels} and OwSGG~\cite{dutta2025open}.
The strong open-set performance indicates that PSGAgent effectively generalizes relational reasoning to unseen predicate categories. 
\begin{table}
  \caption{Recall@100 of each predicate group under the SGDet setting. $\mathcal{S}_{\text{sal}}$ and $\mathcal{S}_{\text{sem}}$ denote the subsets of samples suffering from saliency bias and semantic ambiguity, respectively. Head and Tail indicate splits based on predicate frequency. All denotes the overall performance.}
  \label{tab:predicate_grpo}
  \centering
  \small
  \footnotesize
  \setlength{\tabcolsep}{2.8pt}
  \vspace{-0.1cm}
  \begin{tabular}{l c cc cc c}
    \toprule
    \multirow{2}{*}{\textbf{Model}} &\multirow{2}{*}{\textbf{Params}} & \multicolumn{2}{c}{\textbf{Bottleneck Sets}} & \multicolumn{2}{c}{\textbf{Frequency Sets}} & \multirow{2}{*}{\textbf{All$\uparrow$}} \\
    \cmidrule(lr){3-4} \cmidrule(lr){5-6}
    & & $\mathcal{S}_{\text{sal}}$$\uparrow$ & $\mathcal{S}_{\text{sem}}$$\uparrow$&\textbf{Head}$\uparrow$&\textbf{Tail}$\uparrow$&\\
    \midrule
    Qwen2.5-VL-3B-Instruct            &3B&0.27 & 0.03 &0.58&0.41&0.57\\
    \quad + SFT                       &3B&7.91& 6.58&21.90&5.93&21.67\\
    Qwen2.5-VL-72B-Instruct*          &72B&5.89&3.69&12.39&11.01&12.37\\
    \quad + ZM.T. HL.T.*                &72B&8.98& 6.26&14.50&13.23&14.48\\
    R1-SGG\cite{chen2025compile}      &7B&28.53&24.82&43.72&26.85&43.48\\
    \midrule
    \rowcolor{gray!10} 
    \textbf{PSGAgent (Ours)} & \textbf{3B} & \textbf{40.06} & \textbf{37.45} &\textbf{50.49}& \textbf{34.85} & \textbf{50.27} \\
    \bottomrule
  \end{tabular}
  \vspace{-0.1cm}
\end{table}

\subsubsection{Results under different predicate groups.}
As Table~\ref{tab:predicate_grpo} shows, zero-shot evaluations of the 3B and 72B models reveal severe performance degradation on the saliency bias ($\mathcal{S}_{\text{sal}}$) and semantic ambiguity ($\mathcal{S}_{\text{sem}}$) subsets (e.g., 5.89\%, 3.69\% for 72B model). 
In contrast, their performance across head and tail categories remains relatively balanced (e.g., 12.39\% vs. 11.01\% for the 72B model).
This demonstrates that saliency and semantic ambiguities are fundamental perception bottlenecks inherent to MLLMs and 
the long-tail imbalance is substantially amplified by
task-specific fine-tuning.
Furthermore, equipping the zero-shot 72B model with active visual tools yields larger relative improvements on the $\mathcal{S}_{\text{sal}}$, $\mathcal{S}_{\text{sem}}$, and Tail subsets compared to the overall gain (All).
This trend is further amplified when comparing our PSGAgent against the 3B SFT baseline: it achieves gains of +32.15\% on $\mathcal{S}_{\text{sal}}$, +30.87\% on $\mathcal{S}_{\text{sem}}$, and +28.92\% on Tail, all outpacing the overall improvement (+28.60\% on All).
These disproportionate gains confirm that our dynamic tool-use and training strategy simultaneously alleviate both fundamental perception bottlenecks and dataset-induced long-tail issues.
Finally, our 3B PSGAgent demonstrates comprehensive superiority over the recent 7B R1-SGG model across all metrics, firmly establishing the efficiency and robustness of our agentic framework.
\subsection{Ablation Study}

To better understand the contribution of each component, we conduct ablations on the PSG benchmark under the SGDet setting.


\subsubsection{Effect of perception tools.}
Table~\ref{tab:ablation_sft} evaluates the contributions of the \textit{Region Zoom-In Tool} and the \textit{Visual Highlight Tool}.
 Under SFT-only training, enabling either tool individually improves performance and reduces failure rates, 
 demonstrating that both tools provide complementary perceptual cues.
However, simply combining both tools leads to a slight performance drop. 
This suggests that the imitation-based objective of SFT constrains the model to a fixed output distribution, resulting in suboptimal tool-use strategies. 
After introducing RL, the model learns a more effective tool-use policy, leading to consistent improvements across all metrics. 
The full model, \textbf{PSGAgent}, achieves the best performance with zero failure rate, demonstrating that RL is critical for learning coordinated and
reliable tool invocation strategies.

\subsubsection{Effect of GSPO}
As Table~\ref{tab:ablation_sft} shows, our GSPO variant shows a clear advantage over standard GRPO training, 
bringing an absolute gain of +2.35\% in Recall performance. 
This demonstrates that GSPO is more effective in aligning the agent's reasoning capabilities.

\subsubsection{Effect of reward design.}
Table~\ref{tab:ablation_grpo} studies the impact of different reward signals. 
Removing tool-specific rewards results in clear performance degradation, 
suggesting that generic rewards are insufficient for guiding fine-grained perception behaviors. 
Introducing either strategic highlight or causal zoom rewards individually improves performance, 
as each reward provides targeted supervision for a specific type of perception action.
When both rewards are applied, the model achieves the best performance, indicating that the proposed hierarchical 
reward design provides complementary and synergistic supervision. 
This facilitates more effective credit assignment during RL training, enabling the agent to learn when and how to invoke different tools.

\subsubsection{Effect of M-CoT Dataset.}

As shown in Table \ref{tab:ablation_mcot}, the incorporation of our high-quality 
M-CoT dataset yields a substantial performance leap under the SGDet setting, 
notably boosting mRecall from 13.67\% to 18.51\% (+4.84\%) in the SFT stage and 38.64\% to 42.88\% (+4.24\%) in the GSPO stage. 
Compared to random-sampling baselines in cold-start scenarios, 
M-CoT provides richer, more precise structural priors that enhance open-world generalization.
\subsubsection{Robustness of Mask Segmenter.}
We further examine whether PSGAgent's gains mainly come from a specific external segmenter.
As shown in Table~\ref{tab:highlight_mask_ablation}, the zoom-only variant, which uses no segmentation masks and relies only on the agent's own boxes, already reaches 43.99\% Mean, surpassing the prior MLLM SOTA (SGG-R$^3$~\cite{feng2026sgg}, 42.98\%).
Adding the highlight tool improves Mean by 2.59 points, and replacing Mask2Former with the weaker OneFormer~\cite{oneformer} still achieves 45.24\% Mean, only 1.34 below the default setting.
Moreover, PSGAgent remains stable under synthetic mask noise: even dropping 30\% of masks causes only a 1.99 Mean decrease, retaining 95.7\% of the default performance.
These results indicate that the performance gain is not tied to a particular segmenter, but mainly comes from adaptive tool use and robust relation reasoning.

Overall, these results highlight that effective PSG requires not only access to perception tools but also the ability to strategically coordinate them. 
The combination of M-CoT dataset, tool design and hierarchical reward learning is essential for overcoming the perception bottleneck and achieving robust performance.
\begin{table} 
  \caption{Ablation of tool-use and training strategies on PSG benchmarks under the SGDet setting.}
  \label{tab:ablation_sft}
  \centering
  \footnotesize 
  \setlength{\tabcolsep}{2.8pt} 
  \vspace{-0.1cm}
  \begin{tabular}{l c cc c ccc}
    \toprule
    \multirow{2}{*}{\textbf{Model}} & \textbf{Pre-} & \textbf{Zm.} & \textbf{Hl.} & \textbf{Failure} & \multicolumn{3}{c}{\textbf{PSG Metrics}} \\
    \cmidrule(lr){6-8}
    &\textbf{train} &\textbf{T.} & \textbf{T.} & \textbf{Rate$\downarrow$} & \textbf{AP@50$\uparrow$} & \textbf{Recall$\uparrow$} & \textbf{mRecall$\uparrow$} \\
    \midrule
    
    SFT & -- & \faTimes & \faTimes & 0.82 & 30.10 & 21.67 & 10.37 \\
    SFT & -- & \faTimes & \faCheck & 0.22 & 38.49 & 25.26 & 18.70 \\
    SFT  & -- & \faCheck & \faTimes & 0.20 & 38.95 & 25.47 & 19.74 \\
    SFT & -- & \faCheck & \faCheck & 0.41 & 38.26 & 24.79 &18.51 \\
    GSPO & SFT & \faTimes & \faTimes & 0.01 & 39.73 &39.31& 27.71 \\
    GSPO & SFT & \faTimes & \faCheck & 0.00 & 46.00 & 48.47 & 40.39 \\
    GSPO  & SFT & \faCheck & \faTimes & 0.00 &46.70 & 48.43 & 39.54 \\
    GRPO & SFT & \faCheck & \faCheck & 0.00 & 46.86 & 47.92 & 41.79 \\
    \rowcolor{gray!10}
    \textbf{GSPO} & SFT & \faCheck & \faCheck & \textbf{0.00} & \textbf{49.91} & \textbf{50.27} & \textbf{42.88} \\
    \bottomrule
  \end{tabular}
\end{table}

\begin{table}
  \caption{Ablation of GSPO reward strategies under the SGDet setting. Hl. Rwd and Zm.Rwd denote Highlight and Zoom rewards for tool invocation.}
  \label{tab:ablation_grpo}
  \centering
  \footnotesize 
  \setlength{\tabcolsep}{3pt} 
  \vspace{-0.1cm}
  \begin{tabular}{l c cc c ccc}
    \toprule
    \multirow{2}{*}{\textbf{Model}} & \textbf{Pre-} & \textbf{Zm.} &\textbf{Hl.} &  \textbf{Failure} & \multicolumn{3}{c}{\textbf{PSG Metrics}} \\
    \cmidrule(lr){6-8}
    & \textbf{train} & \textbf{Rwd.} & \textbf{Rwd.} & \textbf{Rate$\downarrow$} & \textbf{AP@50$\uparrow$} & \textbf{Recall$\uparrow$} & \textbf{mRecall$\uparrow$} \\
    \midrule
    
    GSPO & SFT & \faTimes & \faTimes & 0.02 & 34.15 & 43.63 & 35.72 \\
    GSPO & SFT & \faCheck & \faTimes  & 0.04 & 38.92 & 45.72 & 41.17 \\
    GSPO & SFT & \faTimes &\faCheck & 0.03 & 40.07 & 48.49 & 37.83 \\
    \rowcolor{gray!10}
    \textbf{GSPO} & SFT & \faCheck & \faCheck & \textbf{0.00} & \textbf{49.91} & \textbf{50.27} & \textbf{42.88} \\
    \bottomrule
  \end{tabular}
\end{table}

\begin{table}
  \caption{Ablation of the M-CoT dataset under the SGDet setting. 
  w/o M-CoT denotes randomly sampled data used for trajectory synthesis during the initial cold-start phase.}
  \label{tab:ablation_mcot}
  \centering
  \footnotesize 
  \setlength{\tabcolsep}{3pt} 
  \vspace{-0.1cm}
  \begin{tabular}{l c c c ccc}
    \toprule
    \multirow{2}{*}{\textbf{Model}} & \textbf{Pre-} & \textbf{M-CoT}&  \textbf{Failure} & \multicolumn{3}{c}{\textbf{PSG Metrics}} \\
    \cmidrule(lr){5-7}
    & \textbf{train} & \textbf{dataset} & \textbf{Rate$\downarrow$} & \textbf{AP@50$\uparrow$} & \textbf{Recall$\uparrow$} & \textbf{mRecall$\uparrow$} \\
    \midrule
    SFT & -- & \faTimes & 0.42 & 36.98 & 21.51 & 13.67 \\
    SFT & -- & \faCheck & 0.41 & 38.26 & 24.79 & 18.51 \\
    GSPO & SFT & \faTimes & 0.00 & 49.27 & 48.99 & 38.64 \\
    \rowcolor{gray!10}
    \textbf{GSPO} & SFT & \faCheck& \textbf{0.00} & \textbf{49.91} & \textbf{50.27} & \textbf{42.88} \\
    \bottomrule
  \end{tabular}
\end{table}

\begin{table}[t]
  \caption{Ablation of mask sources.}
  \label{tab:highlight_mask_ablation}
  \centering
  \footnotesize 
  \setlength{\tabcolsep}{2pt}
  \vspace{-0.1cm}
  \begin{tabular}{lcccc}
    \toprule
    Setting & \textbf{AP@50} & \textbf{Recall} & \textbf{mRecall} & \textbf{Mean} \\
    \midrule
    \rowcolor{gray!10}
    \multicolumn{5}{l}{\textit{Highlight mask source}} \\
    No Highlight (zoom only)        & 46.70 & 48.43 & 39.54 & 43.99 \\
    OneFormer (weaker segmenter)    & 49.60 & 49.67 & 40.87 & 45.24 \\
    GT panoptic masks (upper bound) & 49.12 & 50.44 & 46.03 & 48.24 \\
    \midrule
    \rowcolor{gray!10}
    \multicolumn{5}{l}{\textit{Mask degradation based on Mask2Former}} \\
    Clean Mask2Former               & --    & 50.27 & 42.88 & 46.58 \\
    + dilation/erosion ($\pm3$px)   & --    & 49.11 & 40.27 & 44.69 \\
    + 30\% masks dropped            & --    & 48.71 & 40.47 & 44.59 \\
    \midrule
    Mask2Former (default)           & 49.91 & 50.27 & 42.88 & \textbf{46.58} \\
    \bottomrule
  \end{tabular}
  \vspace{-0.2cm}
\end{table}

\section{Analysis}

 \begin{figure*}[t] 
  \centering
  \begin{subfigure}[b]{0.49\linewidth}
    \centering
    \includegraphics[width=\linewidth]{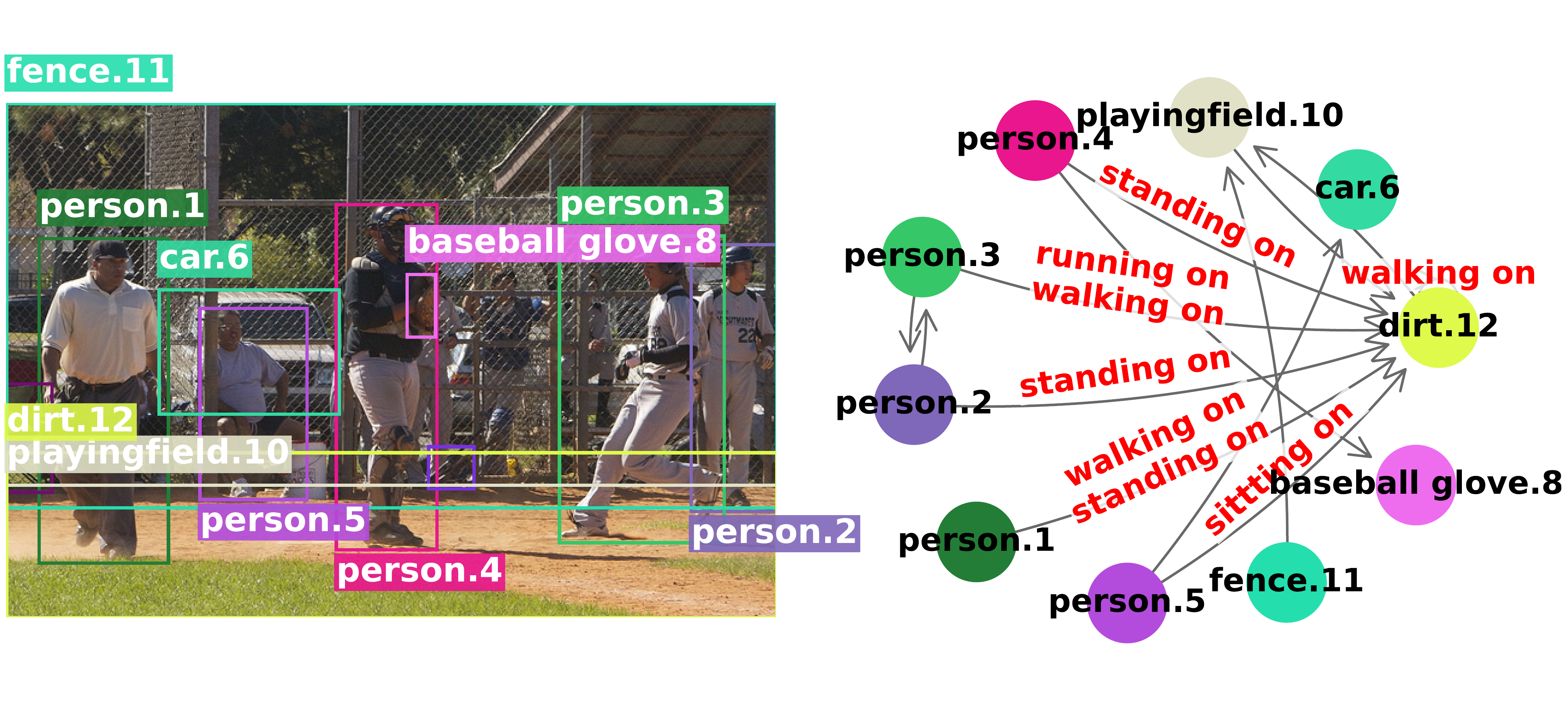} 
    \vspace{-0.8cm}
    \caption{R1-SGG(7B)}
    \label{fig:R1SGG}
  \end{subfigure}
  \hfill 
  \begin{subfigure}[b]{0.49\linewidth}
    \centering
    \includegraphics[width=\linewidth]{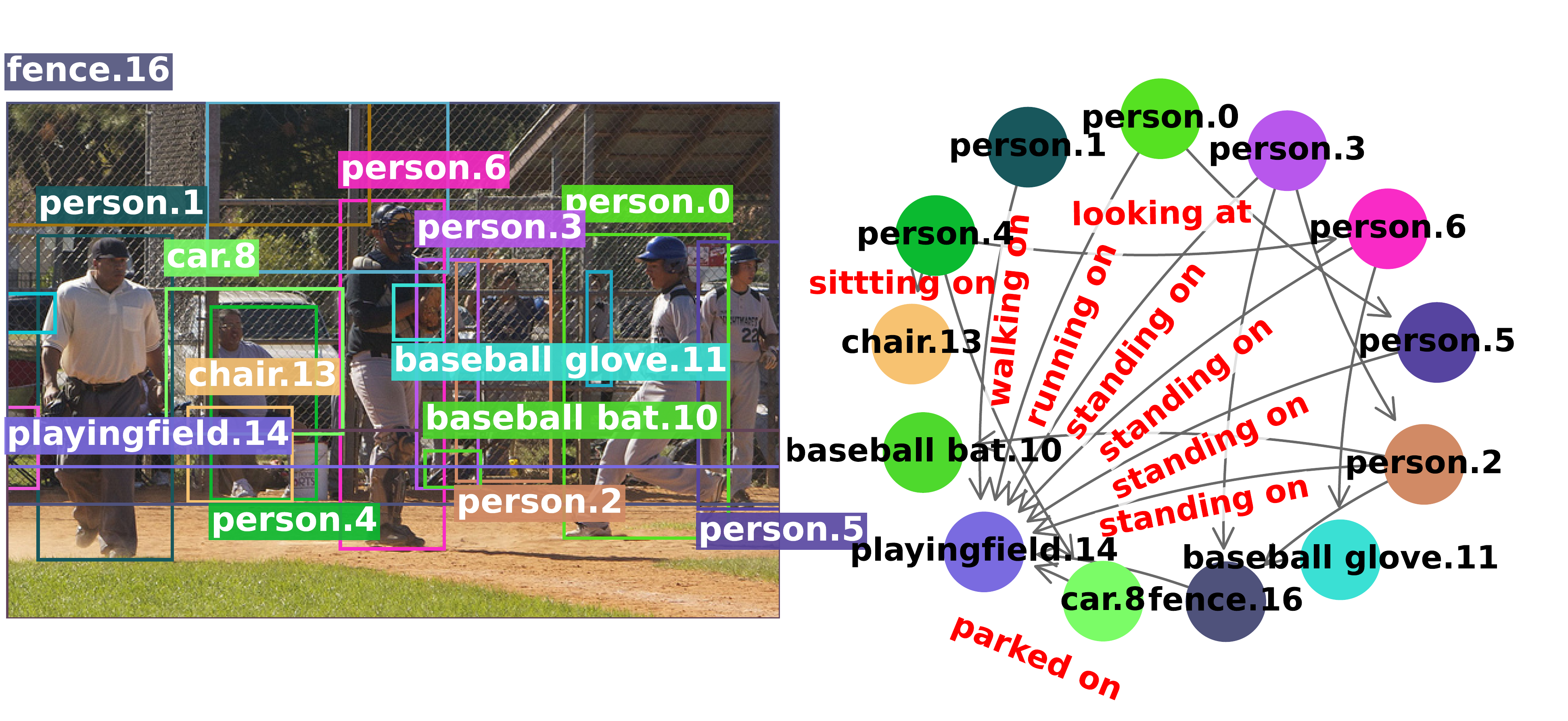} 
    \vspace{-0.8cm}
    \caption{PSGAgent(3B)(Ours)}
    \label{fig:psgagent}
  \end{subfigure}
  \vspace{-0.3cm}
  \caption{Qualitative comparison of PSGAgent with existing methods. (a) R1-SGG (7B); (b) PSGAgent (3B, ours).}
  \label{fig:qualitative_results}
\end{figure*}
\subsection{Qualitative Results}

As illustrated in Figure~\ref{fig:qualitative_results}, 
PSGAgent demonstrates clear advantages in mitigating saliency bias, resolving fine-grained semantic ambiguities, and generating more comprehensive scene graphs.

\subsubsection{Alleviating Saliency Bias and Semantic Ambiguity.} 
Current fixed perception models often suffer from saliency bias.
For instance, R1-SGG~\cite{chen2025compile} misses the less prominent interaction of \textit{\textless person.4 sitting on chair.13\textgreater} and \textit{\textless car.8 parked on playingfield.14\textgreater},
whereas PSGAgent accurately captures these subtle relations. Furthermore, our model effectively resolves semantic ambiguity. 
While competitors like R1-SGG output contradictory predicates for the same subject 
(e.g., predicting a person is simultaneously \textit{walking on} and \textit{standing on} the dirt), 
PSGAgent instead assigns mutually exclusive, precise predicates
(e.g., accurately distinguishing {\textless\textit{person.0 running on}, \textit{person.1 walking on}, and \textit{person.2 standing on} the \textit{playingfield.14}\textgreater}).

\subsubsection{Generating Denser and More Applicable Scene Graphs.} 
Human annotations often suffer from severe missing label issues 
(e.g., 20 objects but only 5 annotated relations in Fig.~\ref{fig:qualitative_results}). 
Instead of overfitting to such sparse ground truth, 
PSGAgent proactively discovers numerous unannotated yet semantically plausible triplets 
(e.g., \textit{\textless person.4, looking at, person.6\textgreater}). This demonstrates its strong open-world reasoning capability, 
yielding more holistic scene graphs.

\subsubsection{Superior Structural Logic and Reduced Hallucination.} Beyond density, PSGAgent demonstrates superior structural logic. 
 While R1-SGG suffers from severe multimodal hallucinations—fabricating illogical relationships 
 like \textit{\textless dirt.12, walking on, dirt.12\textgreater}—
 PSGAgent strictly adheres to semantic and physical constraints, ensuring the generated scene graphs are both rich and logically reliable.

 \begin{figure}[t] 
  \centering
  \begin{subfigure}[b]{0.49\linewidth}
    \centering
    \includegraphics[width=\linewidth]{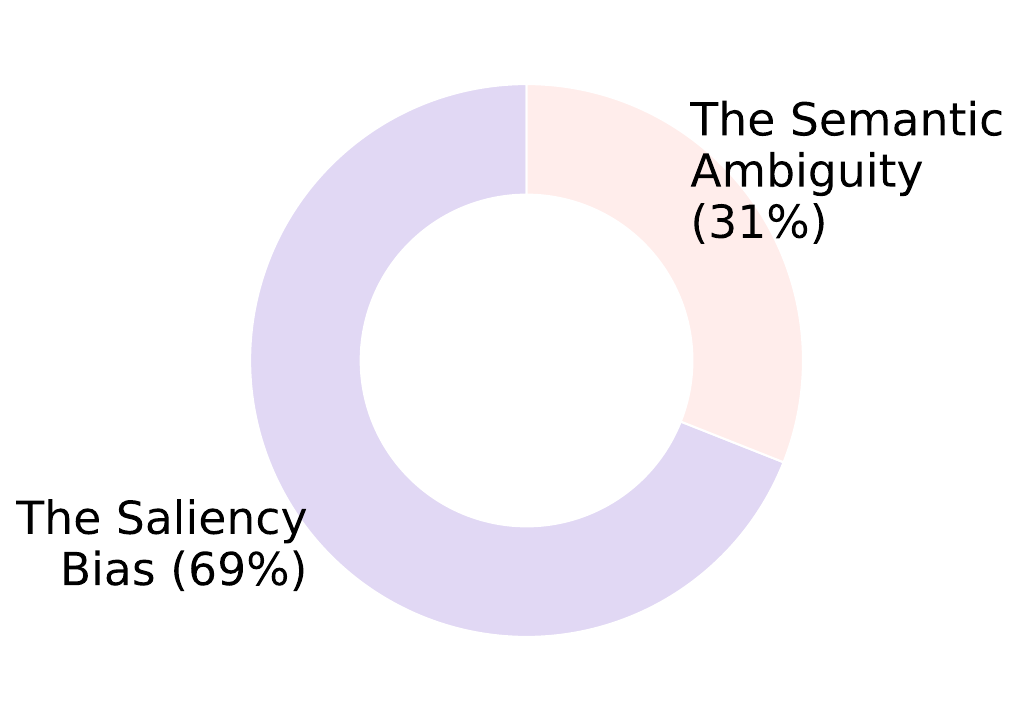} 
    \caption{Distribution of error types}
    \label{fig:error_type}
  \end{subfigure}
  \hfill 
  \begin{subfigure}[b]{0.49\linewidth}
    \centering
    \includegraphics[width=\linewidth]{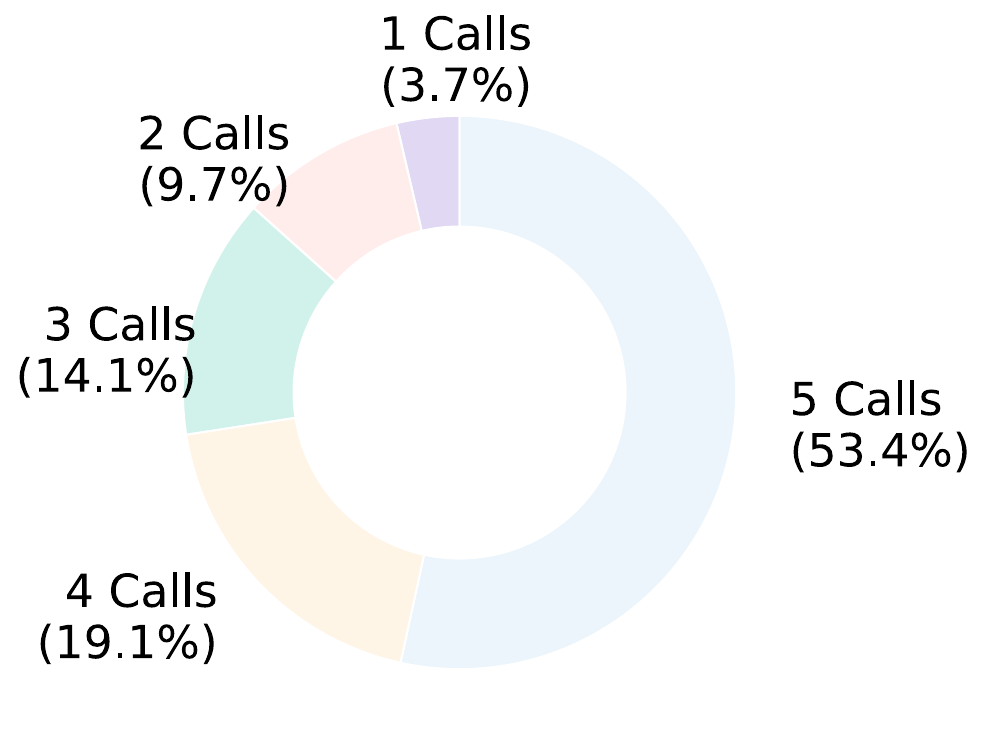} 
    \caption{Rounds of tool calls}
    \label{fig:tool_calls}
  \end{subfigure}
  \vspace{-0.3cm}
  \caption{Analysis of the M-CoT dataset. (a) Classification of error types; (b) Rounds of tool calls.}
  \label{fig:combined_distribution}
\end{figure}

 \begin{figure}[t] 
  \centering
  \begin{subfigure}[b]{0.49\linewidth}
    \centering
    \includegraphics[width=\linewidth]{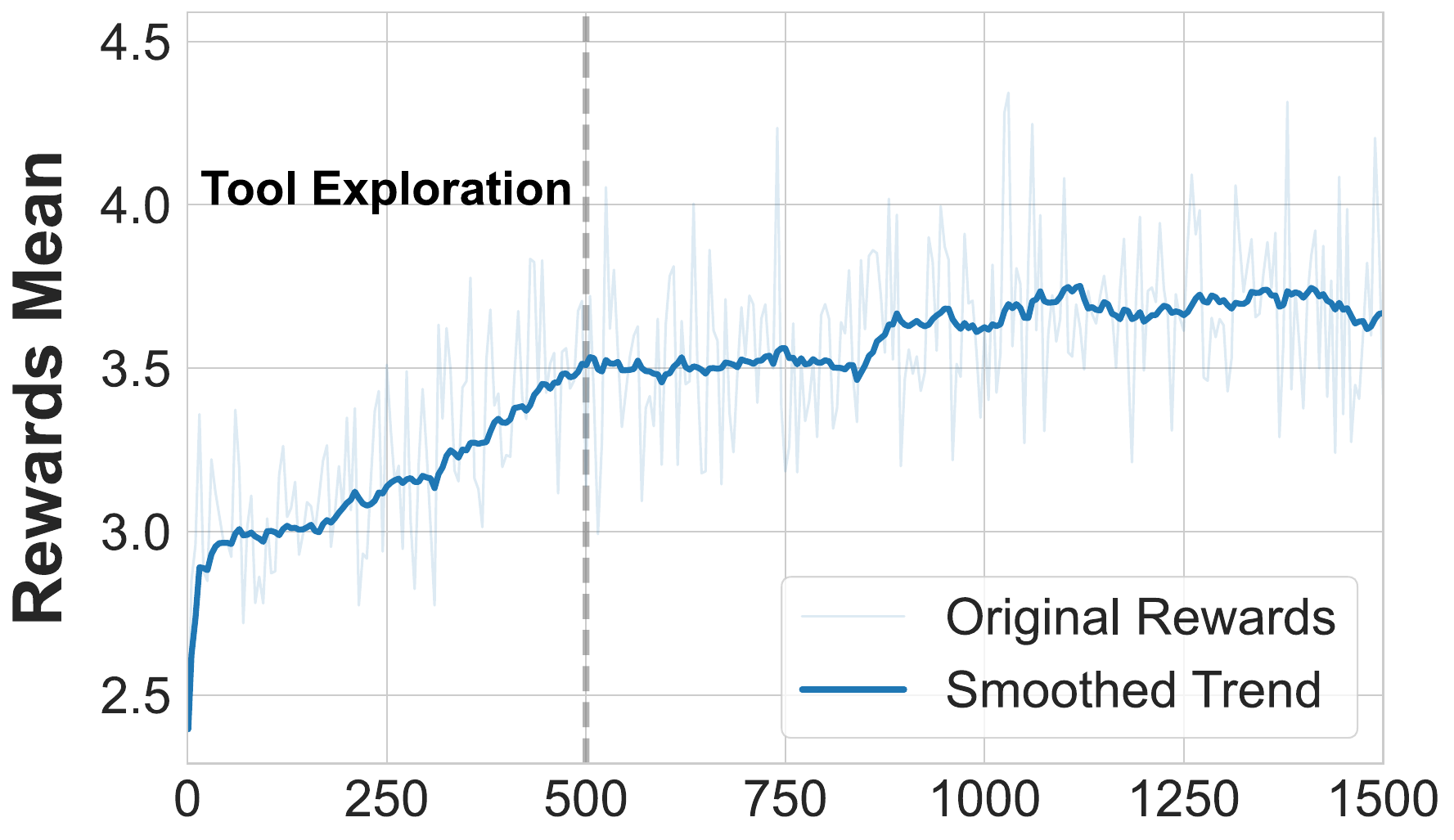} 
    \caption{Rewards Curve}
    \label{fig:train_trewards}
  \end{subfigure}
  \hfill 
  \begin{subfigure}[b]{0.49\linewidth}
    \centering
    \includegraphics[width=\linewidth]{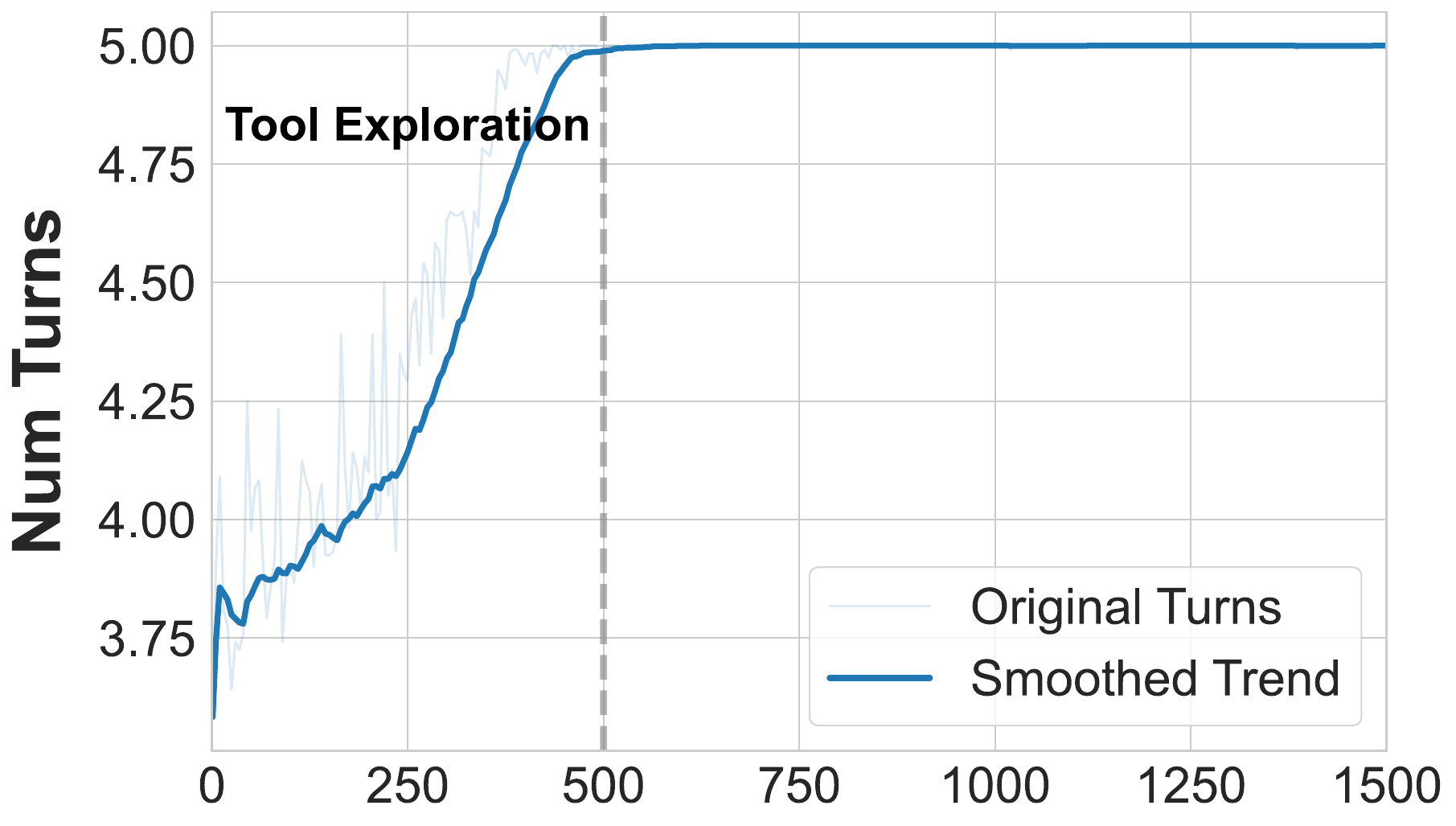} 
    \caption{Tool-Call Curve}
    \label{fig:train_turs}
  \end{subfigure}
  \vspace{-0.3cm} 
  \caption{Training dynamics of GSPO. (a) Rewards curve; (b) Tool-Call Curve.}
  \label{fig:train_distribution}
  \vspace{-0.3cm} 
\end{figure}

\subsection{Analysis of the M-CoT Dataset}
To better understand the challenges in PSG, 
we analyze the error types in the M-CoT dataset (Figure~\ref{fig:error_type}). 
The error distribution is dominated by saliency bias (69\%), followed by semantic ambiguity (31\%).
This confirms that current models primarily struggle with fine-grained relation distinction and 
balanced attention across relations. Additionally, Figure~\ref{fig:tool_calls} reveals that most tasks demand extensive iterative reasoning: 
over 72\% of instances require 4 or 5 rounds of tool calls. This distribution underscores the complexity of the dataset and the necessity of multi-step generation capabilities.
\subsection{Training Dynamics}
Figure~\ref{fig:train_distribution} illustrates the training dynamics of the agent, which consist of two phases.
The early \textit{tool exploration phase} shows a steady reward increase alongside rapid growth in tool calls,
 indicating that the agent is learning to effectively invoke tools. 
Subsequently, both curves stabilize, while the number of tool calls
converges to approximately five per episode.
 This stable reward curve and calling frequency demonstrate that PSGAgent successfully masters a coordinated tool-use strategy for complex reasoning.
\section{Conclusion}
In this paper, we propose PSGAgent, a novel agentic MLLM framework
that addresses the perception bottleneck in existing PSG methods.
By reformulating PSG as a multi-turn reasoning process, our framework empowers the model to dynamically refine its visual observations via a \textit{think--act--observe} loop. 
We equip the agent with the \textit{Region Zoom-In Tool} and the \textit{Visual Highlight Tool} and optimize it through a robust two-stage pipeline: cold-start SFT on a novel error-aware M-CoT dataset, followed by GSPO with hierarchical rewards. 
Extensive experiments demonstrate that PSGAgent achieves SOTA performance on the PSG benchmark, autonomously resolving saliency bias and semantic ambiguity to generate denser, more logical and highly accurate scene graphs. 
Future work will extend this dynamic perception paradigm to Video Scene Graph Generation (VidSGG) and embodied agents.
\begin{acks}
This work was supported by the industry--academia--research
collaboration project between Peking University and
Industrial and Commercial Bank of China Limited
under Grant No.~8430104168.
\end{acks}

\bibliographystyle{ACM-Reference-Format}
\bibliography{main}
\end{document}